\documentclass{llncs}
\usepackage{llncsdoc}
\usepackage{graphicx}

\usepackage{textcomp}
\usepackage{epsfig,graphicx}
\usepackage{xcolor}
\usepackage{amsfonts,amsmath,amssymb}
\usepackage{booktabs}

\usepackage{subfigure,subfloat,algorithm,algorithmic} 

\title{\bf Compression Rate Method for Empirical Science
and Application to Computer Vision}           

\author{
{\bfseries Daniel Burfoot}\\}

\institute{
Independent Researcher \\ 
709 North 7th St., Emmaus, PA 18049 \\
daniel.burfoot@gmail.com
}

\begin{document}

\newcommand{\etal}{\textit{et al.}}

\maketitle                        

\begin{abstract}
This philosophical paper proposes a modified version of the scientific method,
	in which large databases are used instead of experimental observations
	as the necessary empirical ingredient.
This change in the source of the empirical data
	allows the scientific method to be applied to several aspects of physical reality
	that previously resisted systematic interrogation.
Under the new method, 
	scientific theories are compared by instantiating them as compression programs,
	and examining the codelengths they achieve on a database of measurements
	related to a phenomenon of interest.
Because of the impossibility of compressing random data,
	``real world'' data can only be compressed by 
	discovering and exploiting the empirical structure it exhibits.
The method also provides a new way of thinking about two longstanding issues
	in the philosophy of science:
	the problem of induction and the problem of demarcation.

The second part of the paper proposes to reformulate computer vision
	as an empirical science of visual reality,
	by applying the new method to large databases of natural images.
The immediate goal of the proposed reformulation 
	is to repair the chronic difficulties
	in evaluation experienced by the field of computer vision.
The reformulation should bring a wide range of benefits,
	including a substantially increased degree of methodological rigor,
	the ability to justify complex theories without overfitting,
	a scalable evaluation paradigm,
	and the potential to make systematic progress.
A crucial argument is that the change is not especially drastic,
	because most computer vision tasks can be 
	reformulated as specialized image compression techniques.
Finally, a concrete proposal is discussed in which a 
	database is produced by recording from a roadside video camera,
	and compression is achieved by developing a computational understanding
	of the appearance of moving cars.
	
\end{abstract} 

\section{Introduction}

The ultimate ambition of the field of computer vision is 
	to build machines that can see as well as humans.
Despite decades of research, profound theoretical insights,
	sophisticated mathematics, and the arrival of fabulously powerful computers,
	this goal is currently far out of reach.
The position taken in this paper is that the critical failure
	of the field is that is that it does not properly emphasize 
	the role of \textit{empirical science}.

This is not, by itself, an especially interesting or original claim.
Many researchers have lamented the lack of empirical rigor 
	in computer vision~\cite{Haralick:1986,Jain:1991}.
Papers proposing new evaluation methods often contain 
	statements that can be paraphrased as
	``despite the fact that topic X is one of the most heavily researched areas in the field,
	there is still no good way of evaluating the performance of algorithms that perform X''
	~\cite{Scharstein:2002,Bowyer:1999,Estrada:2005}.
This lack of empirical rigor is not, of course,
	due to vision scientists' failure to properly understand
	the scientific method or its importance.
Rather, the problem is that the scientific method,
	in the form that it is currently understood,
	cannot be productively applied to the questions of interest in the field.
It is not obvious, for example, how the scientific method can be used
	to obtain a high-quality segmentation algorithm.
The central contribution of this paper is a refined or adapted version of the 
	scientific method, based on large scale lossless data compression, 
	which can be used for computer vision research.

Section~\ref{sec:develcompr} begins with a discussion of the traditional method and shows,
	through a series of thought experiments,
	how the new method can be derived from it through a series of basically minor revisions.
The new method retains the essential character of the traditional one,
	but is different enough that it can be used to justify several new lines of inquiry.
In particular, the new method uses large databases instead of controlled experiments
	as the key empirical ingredients that are necessary to 
	test, refine, and potentially falsify a theory.
Since it is relatively easy to obtain large databases of natural images,
	the new method can be applied to these databases 
	to guide a search for theories of \textit{visual reality}.
	
A critical aspect of the new method is its emphasis on empirical science.
The No Free Lunch theorem of data compression indicates that
	compression can only be achieved for some bit strings
	at the price of inflating others.
This implies that the only way to compress a large database of natural images
	is by discovering and exploiting the empirical regularities
	contained in the images.
In other words,
	every compressor contains an implicit assertion or hypothesis 
	about the data to which it will be applied,
	and if it succeeds in achieving compression, 
	that fact provides empirical evidence that the hypothesis is correct.
	
A related point concerns the problem of demarcation:
	how can one tell the difference between science and pseudoscience?
Popper suggested to solve this problem by accepting a theory as scientific 
	if and only if it makes explicit anti-predictions,
	and thereby exposes itself to falsification if the prohibited event
	actually occurs as an experimental result~\cite{Popper:1959}.
The compression principle suggests an analogous requirement,
	since the only way to achieve compression is by reassigning probability away
	from some outcomes and towards others.
Here the demarcation problem is solved by identifying 
	as unscientific those theories which fail to perform such a reassignment
	(such theories cannot achieve compression).
Furthermore, if the theory performs a probability reassignment,
	but does so in a way that does not correspond to empirical reality,
	it will end up inflating the database and is therefore falsified.

Another philosophically significant aspect of the new method is its
	use of large (``vast'') datasets to overcome the problem of induction.
This argument relates to the concept of Kolmogorov complexity~\cite{Vitanyi:1997}.	
The Kolmogorov complexity of a large and complex bit string is
	an absolute quantity that does not depend on an observer's prior belief
	(encoded in his choice of Turing machine or programming language).
If one researcher claims that a new theory achieves a certain compression rate on a large database,
	this claim can be verified by all observers regardless of their prior beliefs.
Thus, in this ``vast data regime'', statistical inference becomes objective.

The second part of the paper, contained in Section~\ref{sec:comprvsion},
	shows how the new method can be applied to computer vision research.
First, Section~\ref{sec:mthoddiffc} provides a brief critical review 
	of current evaluation methods.
While some progress has been made recently in 
	this area~\cite{Scharstein:2002,Bowyer:1999,Martin:2001,Feifei:2007}
	it is clear that deep conceptual issues remain~\cite{Ponce:2006,Unnikrishnan:2007}.
	
Next, Section~\ref{sec:advantages} describes the various advantages
	that can be achieved by formulating
	computer vision in terms of large scale image compression.
The most obvious advantage is methodological rigor:
	a compressor is invoked on a database, 
	the encoded file size is recorded,
	and then the decoded version is checked to make sure it matches
	the original exactly.
The evaluation principle is also scalable:
	a single unlabeled database can be used to 
	justify and evaluate the performance of many different techniques.
Since the new method mandates the use of large unlabeled databases,
	highly complex models can be developed without overfitting.
Finally, by conducting a determined search for the package of methods 
	that achieve the shortest possible codelength,
	the field can make systematic progress.
Thus, by changing the nature of the questions it considers,
	the field can realize enormous methodological advantages.

A crucial argument, contained in Section~\ref{sec:equivalence},
	is that the shift from the traditional formulation of vision
	to the vision-as-compression approach
	is not really very drastic.
Nearly all computer vision tasks can be easily reformulated
	as special techniques for image compression.
For example, image segmentation can be viewed as a special
	way of compressing an image by separating the pixels into homogeneous regions,
	so that bits can be saved by encoding the pixels with a 
	region-specific model~\cite{Leclerc:1989,Zhu:1996}.
The stereo matching problem can be reformulated as a special way of compressing
	a stereo pair by using the first image, plus a disparity function,
	to predict the pixels in the second image~\cite{Mumford:1994}.
	
To make the ideas of this paper as concrete as possible,
	Section~\ref{sec:cncrtprpsl} gives one simple proposal.
Here, a camera is set up next to a highway,
	and is used to obtain a large video database.
Assuming the background is mostly static,
	the major source of variation in the video
	will be the passing cars.
Thus, in order to achieve good compression rates for these images,
	it will be necessary to develop a sophisticated computational understanding
	of the appearance of automobiles.
This understanding can be built up in several levels,
	with each level allowing a better compression rate than the previous one.
The first level would exploit the fact that cars are rigid bodies
	obeying Newtonian laws of motion.
Higher levels would exploit characteristic visual properties of automobiles,
	such as the appearance of the wheels and the windshields. 

\section{Development of Compression Rate Method}
\label{sec:develcompr}

\subsection{Traditional Scientific Method}

This paper proposes a refined version of the scientific method
	that is more directly applicable to the problems of interest
	in computer vision.
Before doing so,
	it is worth briefly examining the traditional method
	and the circumstances in which it can be applied.
The scientific method is not an exact procedure,
	but a qualitative statement of it goes roughly as follows:

\begin{enumerate}
\item Observe a natural phenomenon.
\item Develop a theory of that phenomenon.
\item Use the theory to make a prediction.
\item Test the prediction experimentally.
\end{enumerate}

A full discussion of the philosophical significance
	of the scientific method is beyond the scope of this paper,
	but some brief remarks are in order.
The power of the scientific method is in the way it links
	theory with experimental observation;
	either one of these alone is worthless.
The long checkered intellectual history of humanity clearly shows
	how rapidly pure theoretical speculation goes astray
	when it is not tightly constrained by an external guiding force.
Pure experimental investigation, in contrast, 
	is of limited value because of the vast number of possible configurations of objects.
To make predictions solely on the basis of experimental data,	
	it would be necessary to exhaustively test each configuration.

As articulated in the above list, 
	the goal of the method appears to be the verification of a single theory.
This is a bit misleading;
	in reality the goal of the method is to facilitate selection between a potentially large number of candidate theories.
Given two competing theories of a particular phenomenon,
	the researcher identifies some experimental configuration where 
	the theories make incompatible predictions and then 
	performs the experiment using the indicated configuration.
The theory whose predictions fail to match the experimental prediction
	is discarded in favor of its rival.
But even this view of science as a process of weeding out imperfect
	theories in order to find the perfect one is somewhat inaccurate.
Most physicists will admit or disclaim that even their most refined theories
	are mere approximations,
	though they are spectacularly accurate approximations.
The scientific method can therefore be understood as a technique 
	for using empirical observations 
	to find the best predictive approximation 
	from a large pool of candidates.

A core component of the traditional scientific method 
	is the use of controlled experiments. 
To control an experiment means essentially to simplify it.
To determine the effect of a certain factor,
	one sets up two experimental configurations
	which are exactly the same except for the presence or absence of the factor.
If the experimental outcomes are different,
	then it can be inferred that this disparity is due to the special factor.

In some fields of scientific inquiry, however,
	it is impossible or meaningless to conduct controlled experiments.
No two people are identical in all respects,
	so clinical trials for new drugs,
	in which the human subject is part of the experimental configuration,
	can never be truly controlled.
The best that medical researchers can do is to attempt to ensure
	that the experimental factor does not systematically correlate
	with other factors that may affect the outcome.
This is done by selecting at random which patients will receive the new treatment.
This method has obvious limitations, however,
	which cause deep problems in the medical literature~\cite{Ioannidis:2006}.
It is similarly difficult to apply the traditional scientific method
	to answer questions arising in the field of macroeconomics.
No political leader would ever agree to a proposal in which her country's 
	economy was to be used as an experimental test subject.
In lieu of controlled experiments, 
	economists attempt to test their theories based on the 
	outcomes of so-called historical experiments,
	where two originally similar countries implemented different economic policies.
	
A similar breakdown of the traditional method occurs in computer vision.
Controlled vision experiments can be conducted,
	but are of very little interest.
The physical laws of reflection and optics that govern the image formation 
	process are well understood already.
Clearly if the same camera is used to photograph an identical scene
	twice under constant lighting conditions, 
	the obtained images will be identical or very nearly so.
And a deterministic computer vision algorithm will always produce the same
	result when applied to two identical images.
It is not clear, therefore, how to use the traditional method 
	to approach the problems of interest in computer vision,
	which include tasks like image segmentation and edge detection.
	
\subsection{Sophie's Adventures}

By making a series of minor modifications to the traditional scientific method,
	this paper develops a refined version that
	is directly applicable to the problems of interest in computer vision.
These modifications are illustrated through a series 
	of thought experiments relating to a fictional character named Sophie.

\subsubsection{The Shaman}

Sophie is a assistant professor of physics at a large American state university.
She finds this job vexing for several reasons,
	one of which is that she has been chosen by the department 
	to teach a physics class intended for students majoring in the humanities,
	for whom it serves to fill a breadth requirement. 
The students in this class, who major in subjects like literature,
	religious studies, and philosophy, tend to be intelligent
	but also querulous and somewhat disdainful of the ``merely technical'' 
	intellectual achievements of physics.

In the current semester she has become aware of the presence
	in her class of a discalced student with a large beard and often bloodshot eyes.
This student is surrounded by an entourage of similarly odd-looking followers.
Sophie is on good terms with some of the more serious students in the class,
	and in conversation with them has found out that the odd student
	is attempting to start a new naturalistic religious movement and refers to himself as a ``shaman''.

One day while delivering a simple lecture on Newtonian mechanics,
	she is surprised when the shaman raises his hand and claims
	that physics is a propagandistic hoax designed by the elites
	as a way to control the population.
Sophie blinks several times, 
	and then responds that physics can't be a hoax because it makes
	real-world predictions that can be verified by independent observers.
The shaman counters by claiming that the so-called ``predictions'' made by physics 
	are in fact trivialities, and that he can obtain better forecasts by communing with the spirit world.
He then proceeds to challenge Sophie to a predictive duel,
	in which the two of them will make forecasts regarding the outcome of a simple experiment,
	the winner being decided based on the accuracy of the forecasts.
Sophie is taken aback by this but,
	hoping that by proving the shaman wrong she can break the spell he has cast on some of the other students,
	agrees to the challenge.
	
During the next class, Sophie sets up the following experiment. 
She uses a spring mechanism to launch a ball into the air at an angle $\theta$.
The launch mechanism allows her to set the initial velocity of the ball 
	to a value of $v_{i}$.
She chooses as a predictive test the problem of predicting the time $t_{f}$
	that the ball will fall back to the ground after being launched at $t_{i}=0$.
Using a trivial Newtonian calculation she concludes that $t_{f} = 2 g^{-1} v_{i} \sin(\theta)$,
	sets $v_{i}$ and $\theta$ to give a value of $t_{f} = 2$ seconds,
	and announces her prediction to the class.
She then asks the shaman for his prediction.
The shaman declares that he must consult with the wind spirits,	
	and then spends a couple of minutes chanting and muttering.
Then, dramatically flaring open his eyes as if to signify a moment of revelation,
	he grabs a piece of paper, writes his prediction on it,
	and then hands it to another student. 
Sophie suspects some kind of trick, 
	but is too exasperated to investigate and so launches the ball into the air.
The ball is equipped with an electronic timer that starts and stops when an impact is detected,
	and so the number registered in the timer is just the time of flight $t_{f}$.
A student picks up the ball and reports that the result is $t_{f} = 2.134$.
The shaman gives a gleeful laugh,
	and the student holding his written prediction hands it to Sophie.
On the paper is written $1 < t_{f} < 30$.
The shaman declares victory:
	his prediction turned out to be correct,
	while Sophie's was incorrect (it was off by $0.134$ seconds).

To counter the shaman's claim and because it was on the syllabus anyway,
	in the next class Sophie begins a discussion of probability theory.
She goes over the basic ideas,
	and then connects them to the experimental prediction made about the ball.
She points out that technically, the Newtonian prediction $t_{f}=2$ 
	is not an assertion about the exact value of the outcome.
Rather it should be interpreted as the mean of a probability distribution
	describing possible outcomes.
For example, one might use a normal distribution with mean $\mu=t_{f}=2$
	and $\sigma=.3$.
The reason the shaman superficially seemed to win the contest is that he gave  
	a probability distribution while Sophie gave a point prediction;
	these two types of forecast are not really comparable.
In the light of probability theory, 
	the reason to prefer the Newtonian prediction above the shamanic one,
	is that it assigns a higher probability to the outcome that actually occurred.
Now, plausibly, if only a single trial is used then the Newtonian theory might simply have gotten lucky, 
	so the reasonable thing to do is combine the results over many trials, 
	by multiplying the probabilities together.
Therefore the real reason to prefer the Newtonian theory to the shamanic theory
	is that:
	
\begin{equation}
\label{eq:newtnshamn}
\prod_{k} P_{newton}(t_{f,k}) > \prod_{k} P_{shaman}(t_{f,k})
\end{equation}

Where the $k$ index runs over many trials of the experiment.
Sophie then shows how the Newtonian probability predictions are both more \textit{confident}
	and more \textit{correct} than the shamanic predictions.
The Newtonian predictions assign a very large amount of probability
	to the region around the outcome $t_{f}=2$,
	and in fact it turns out that almost all of the real data outcomes fall in this range.	 
In contrast, the shamanic prediction assigns a relatively small amount 
	of probability to the $t_{f}=2$ region, 
	because he has predicted a very wide interval ($1 < t_{f} < 30$).
Thus while the shamanic prediction is correct, it is not very confident.
The Newtonian prediction is correct and highly confident,
	and so it should be prefered.
	
Sophie tries to emphasize that the Newtonian probability prediction $P_{newton}$
	only works well for the \textit{real} data.
Because of the requirement that probability distributions be normalized,
	the Newtonian theory can only achieve superior high performance
	by reassigning probability towards the region around $t_{f} = 2$
	and away from other regions.
A theory that does not perform this kind of reassignment 	
	cannot achieve superior high performance.
	
Sophie recalls that some of the students are studying computer science
	and for their benefit points out the following.
The famous Shannon equation $L(x) = -\log_{2} P(x)$ 
	governs the relationship between the probability of an outcome
	and the length of the optimal code that should be used to represent it.
Therefore, 
	given a large data file containing the results of many trials of the ballistic motion experiment,
	the two predictions (Newtonian and shamanic) can both be used to 
	build specialized programs to compress the data file.
Using the Shannon equation, the above inequality can be rewritten as follows:

\begin{equation}
\sum_{k} L_{newton}(t_{f,k}) < \sum_{k} L_{shaman}(t_{f,k})
\end{equation}

This inequality indicates an alternative criterion
	that can be used to decide between two rival theories. 
Given a data file recording measurements related to a phenomenon of interest, 
	a scientific theory can be used to write a compression program
	that will shrink the file to a small size.
Given two rival theories of the same phenomenon,
	one invokes the corresponding compressors on a shared benchmark data set,
	and prefers the theory that achieves a smaller encoded file size.
This criterion is equivalent to the probability-based one,
	but has the advantage of being more tangible,
	since the quantities of interest are file lengths instead of probabilities.

%
%
%
	
\subsubsection{The Dead Experimentalist}

Sophie is a theoretical physicist and,
	upon taking up her position as assistant professor,
	began a collaboration with a brilliant experimental physicist 
	who had been working at the university for some time.
The experimentalist had previously completed the development 
	of an advanced apparatus that allowed the investigation of an
	exotic new kind of quantum phenomenon.
Using data obtained from the new system,	
	Sophie made rapid progress in developing a mathematical 
	theory of the phenomenon.
Tragically, 
	just before Sophie was able complete her theory,
	the experimentalist was killed in a laboratory explosion
	that also destroyed the special apparatus.
After grieving for a couple of months,
	Sophie decided that the best way to honor her friend's memory
	would be to bring the research they had been working on to a successful conclusion.
	
Unfortunately, there is a critical problem with Sophie's plan.
The experimental apparatus had been completely destroyed,
	and Sophie's late partner was the only person in the world who could have rebuilt it.
He had run many trials of the system before his death,
	so Sophie had a quite large quantity of data.
But she had no way of generating any new data.
Thus, no matter how beautiful and perfect her theory might be,
	she had no way of testing it by making predictions.
	
One day while thinking about the problem 
	Sophie recalls the incident with the shaman.
She remembers the point she had made for the benefit of the software engineers,
	about how a scientific theory could be used to compress a real world data set
	to a very small size.
Inspired, she decides to apply the data compression principle
	as a way of testing her theory. 
She immediately returns to her office and spends 
	the next several weeks	writing Matlab code,
	converting her theory into a compression algorithm.
The resulting compressor is successful:
	it shrinks the corpus of experimental data from an initial size
	of $8.7 \cdot 10^{11}$ bits to an encoded size of $3.3 \cdot 10^{9}$ bits. 
Satisfied, Sophie writes up the theory, 
	and submits it to a well-known physics journal.

The journal editors like the theory, 
	but are a bit skeptical of the compression based method for testing the theory.
Sophie argues that if the theory becomes widely known,
	one of the other experts in the field
	will develop a similar apparatus, 
	which can then be used to test the theory in the traditional way.
She also offers to release the experimental data,
	so that other researchers can test their own theories
	using the same compression principle.
Finally she promises to release the source code of her program,
	to allow external verification of the compression result.
These arguments finally convince the journal editors to accept the paper.
	
\subsubsection{The Upstart Theory}

After all the mathematics, software development,
	prose revisions, and persuasion necessary to complete her theory
	and have the paper accepted,
	Sophie decides to reward herself by living the good life for a while.
She is confident that her theory is essentially correct,
	and will eventually be recognized as correct by her colleagues.
So she spends her time reading novels and hanging out in coffee shops with her friends.

A couple of months later, however,
	she receives an unpleasant shock in the form of an email from a colleague
	which is phrased in consolatory language,
	but does not contain any clue as to why such language might be in order.
After some investigation she finds out that 
	a new paper has been published about the same quantum phenomenon of interest to Sophie.
The paper proposes a alternative theory of the phenomenon which 
	bears no resemblance whatever to Sophie's.
Furthermore, 
	the paper reports a better compression rate than was achieved by Sophie,
	on the database that she released.
	
Sophie reads the new paper and quickly realizes that it is worthless.
The theory depends on the introduction of a large number of additional parameters,
	the values of which must be obtained from the data itself.
In fact, a substantial portion of the paper involves
	a description of a statistical algorithm
	that estimates optimal parameter values from the data.
In spite of these aesthetic flaws,
	she finds that many of her colleagues are quite taken
	with the new paper and some consider it to be ``next big thing''.
	
Sophie sends a message to the journal editors describing in detail
	what she sees as the many flaws of the upstart paper.
She emphasizes the asthetic weakness of the new theory,
	which requires tens of thousands of new parameters.
The editors express sympathy,
	but point out that the new theory outperforms Sophie's theory
	using the performance metric she herself proposed.
The beauty of a theory is important,
	but its correctness is ultimately more important.

Somewhat discouraged, Sophie sends a polite email to the authors of the new paper,
	congratulating them on their result and asking to see their source code.
Their response, which arrives a week later,
	contains a vague excuse about how the source code is not properly documented 
	and relies on proprietary third party libraries.
Annoyed, Sophie contacts the journal editors again and asks them 
	for the program they used to verify the compression result.
They reply with a link to a binary version of the program.

When Sophie clicks on the link to download the program,
	she is annoyed to find it has a size of 800 megabytes.
But her annoyance is quickly transformed into enlightenment,
	as she realizes what happened, and that her previous philosophy contained a serious flaw.
The upstart theory is not better than hers;
	it has only succeeded in reducing the size of the encoded data 
	by dramatically increasing the size of the compressor.
Indeed, when dealing with specialized compressors,
	the distinction between ``program'' and ``encoded data'' becomes almost irrelevant.
The critical number is not the size of the compressed file,
	but the net size of the encoded data plus the compressor itself.
 
Sophie writes a response to the new paper 
	which describes the refined compression rate principle.
She begins the paper by reiterating the unfortunate circumstances
	which forced her to appeal to the principle,
	and expressing the hope that someday an experimental group
	will rebuild the apparatus developed by her late partner,
	so that the experimental predictions made by the two theories 
	can be properly tested.
Until that day arrives, standard scientific practice does not permit a decisive declaration
	of theoretical success.
But surely there is \textit{some} theoretical statement that can be made in the meantime,
	given the large amount of data available.
Sophie's proposal is that the goal should be to find the theory 
	that has the highest probability of predicting a new data set,
	when it can finally be obtained.
If the theories are very simple in comparison to the data being modeled, 
	then the size of the encoded data file is a good way of 
	choosing the best theory.
But if the theories are complex,
	then there is a risk of \textit{overfitting} the data.
To guard against overfitting complex theories must be penalized;	
	a simple way to do this is to take into account
	the codelength required for the compressor itself.
The length of Sophie's compressor was negligible,	
	so the net score of her theory is just the codelength
	of the encoded data file: $3.3 \cdot 10^{9}$ bits.
The rival theory achieved a smaller size of $2.1 \cdot 10^{9}$ for the encoded data file,
	but required a compressor of $6.7 \cdot 10^{9}$ bits to do so,
	giving a total score of $8.8 \cdot 10^{9}$ bits.
Since Sophie's net score is lower, her theory should be prefered.
	
\subsection{Compression Rate Method}

In the course of the thought experiments discussed above,
	the protagonist Sophie articulated what can be considered
	a new method of scientific inquiry.
This procedure is called the Compression Rate Method (CRM)
	in the subsequent development
	and consists of the following steps: 

\begin{enumerate}
\item Obtain a ``vast'' database $T$ relating to a phenomenon of interest.
\item Develop a theory of that phenomenon and 
	use it to construct a compressor.
\item Score the theory by calculating the sum of the
	length of the compressor and the length of the encoded version of $T$.
\item Prefer the new theory to a rival if it achieves a lower score.
\end{enumerate}

Above it was argued that the power of the traditional scientific method
	depended on two core philosophical elements;
	these elements are retained in the new method. 
Both methods employ theoretical speculation,
	but this theorizing is guided, constrained, and verified by empirical data.
And both methods provide a principle for making decisive comparisons between theories, 
	enabling the scientific community to carry out 
	an efficient and systematic search for the best possible theory. 	
	
In spite of this core similarity,
	there is a crucial difference between the traditional method and the CRM.
This difference involves the form
	of the empirical data used to verify a theory.
Physics and related fields depend strongly 
	on the use of controlled experiments to produce the necessary empirical data.
In physics, whenever a new theory of a phenomenon is proposed,
	it must agree with the current theory 
	in a very wide range of scenarios, 
	since the current theory has presumably been shown to 
	make correct predictions on a large number of different configurations.
Controlled experiments are therefore necessary to bring the 
	differing predictions indicated by the two theories
	into the starkest possible relief.
In contrast, the new method
	relies on large quantities of raw data.
Instead of point predictions regarding the outcome of controlled experiments,
	theories justified by the new method are required
	to make probability predictions for potentially complex, high-dimensional data.
It no longer makes sense to say that the current theory makes definitively ``correct'' 
	predictions for some standard configurations; 
	rather, the predictions are just ``good''. 
To supplant the current theory, a new theory need only make better predictions on average.

Because of the relaxation of the requirement for controlled experiments,
	the new method immediately justifies several new lines of scientific inquiry.
The type of science produced by CRM-style inquiry depends on the 
	type of data contained in the target database $T$.
If $T$ contains data related to the outcomes of physical experiments,
	then physical theories will be necessary to compress it.
If $T$ contains information related to interest rates, 
	house prices, global trade flows, and so on,
	then economic theories will be necessary to compress it.
But the most obvious choice for $T$ 
	is simply an enormous image database,
	such as those built by photo sharing web sites like Flickr.com.
In order to compress such a database one must develop 
	theories of \textit{visual reality}.

In addition to justifying several new lines of scientific inquiry,
	the new method also bears a much stronger resemblance
	to human learning.
The example of human learning shows that, 
	if science is defined as any process by which reliable predictions about the real world are obtained,
	then the traditional method is incomplete.
Human children are able to obtain highly refined models of various phenomena
	that support, in some cases, amazingly accurate predictions.
But children rarely engage in controlled experimentation,
	and in spite of this are able to learn complex skills 
	like basketball and object recognition.
Instead of controlled experimentation, 
	children constantly engage in high-bandwidth interaction with empirical reality
	(also known as ``play''),
	and it is clear that this activity produces a vast amount of 
	raw sensorimotor data from which sophisticated models of various 
	phenomena can be extracted.
The form of the input data assumed in the new method is thus much closer
	to the natural setting of the learning problem as experienced by human children 
	than either the limited size labeled data sets normally used in supervised learning research
	or the controlled experimental data normally used in physics.

\subsubsection{Vast Data and Intersubjective Verifiability}

The history of statistics has observed a major philosophical struggle
	between two schools of thought,
	which advocate opposing perspectives regarding the meaning
	and justification of statistical inference. 
On one side are the Bayesians, 
	who perform inference by choosing a prior distribution 
	and updating it in response to evidence.
On the other side are the frequentists,
	who object to the use of priors.
The frequentist critique of Bayesian methods has two parts.
The first and more accusatory part is that
	Bayesians can (and sometimes do) manipulate the results of their analyses
	by picking convenient priors.
Since statistical analyses often relate to politically sensitive topics such 
	as global warming or the efficacy of new drugs,
	this critique is obviously quite incendiary.
The second and more abstract point
	is that even if the Bayesians do not engage in active intellectual dishonesty,
	there is still no objective way to select a prior.
Since two researchers who start from different priors
	will reach different conclusions
	in spite of observing the same evidence,
	this appears to render statistical inference completely subjective.
		
Any mapping $S(X)$ from data sets $X$ to prefix-free codes
	that is not trivially suboptimal
	implicitly defines a probability distribution $P(X) = 2^{-|S(X)|}$ over data sets.
Since every compressor must implement such a mapping,
	it would appear that CRM research makes the same philosophical
	commitment to the use of prior distributions as do the Bayesians.	
But in fact there is an important difference,
	due to the model complexity penalty term used in the CRM
	(i.e. the length of compressor itself).
By taking this term into account, 
	the prior is now determined essentially by the choice of programming language
	used to write the compressor.
Therefore, CRM researchers can provide an invincible counterargument 
	to the first part of the frequentist critique,
	simply by fixing the programming language in advance and using 
	the same one for all analyses.	

An analysis of the concept of \textit{intersubjective verifiability}
	shows that the compression principle also provides
	a strong defense against the second component of the frequentist critique.
The defense begins by noting that not even traditional science is truly objective. 
Given the same physical evidence about the motions of the planets, 
	there is no guarantee whatever that all observers will draw the same conclusions:
	some may conclude that the Earth rotates around the Sun, others may conclude the opposite.
Instead, traditional science has a property called intersubjective verifiability.
Every scientist has a unique set of formative educational experiences,
	cognitive biases, prefered methods, and analogical schemas with which 
	to interpret the world.
Intersubjective verifiability means that in spite of this intellectual diversity,
	the correctness of a new theory in a given field can be verified
	by every scientist of that field on the basis of known evidence.
	
The concept of universal computation
	provides the same guarantee of intersubjective verifiability 
	to theories justified by sufficiently large quantities of data.
To see this, 
	imagine a research subfield which has established a database $T$
	as its target for CRM-style investigation.
The subfield makes slow but steady progress for several years.
Then, out of the blue,
	an unemployed autodidact from a rural village in India,
	of whom no one has ever heard,
	appears with a bold new theory.
He claims that his theory, instantiated in a program $P_{A}$,
	achieves a compression rate which is dramatically superior
	to the current best published results
	(here the arbitrary distinction between the program and the encoded data is dropped,
	so his codelength is just $|P_{A}|$).
However, among his other eccentricities, 
	this gentleman uses a programming language he himself developed,
	which corresponds to a Turing machine $A$.
Now, the other researchers of the field are well-meaning but skeptical,
	since all the previously published results used a standard language 
	corresponding to a Turing machine $B$.
But in fact it is easy for the Indian maverick to produce a compressor
	that will run on $B$: 
	he simply appends $P_{A}$ to a simulator program $S_{AB}$,
	that simulates $A$ when run on $B$.
The length of the new compressor is $|P_{B}| = |P_{A}| + |S_{AB}|$,
	and all of the other researchers can confirm this.
Now, assuming the data set $T$ is large and complex enough
	so that $|P_{A}| \gg |S_{AB}|$,
	then the codelength of the modified version is 
	effectively the same as the original: $|P_{B}| \approx |P_{A}|$.
This illustrates the meaning of the word ``vast'' for the purposes of CRM research:
	a vast database is one whose size is far larger
	than the characteristic length of a Turing machine simulator program.
In the vast data regime,
	statistical inference attains the same quality of near-objectivity
	that is possessed by traditional science.

\subsubsection{Data Compression as Empirical Science}
	
The following theorem is well known in data compression.
Let $C$ be a program that losslessly compresses bit strings $x$,
	assigning each string to a new code with length $L_{C}(x)$.
Let $U_{N}(x)$ be the uniform distribution over $N$-bit strings.
Then the following bound holds for all compression programs $C$:

\begin{equation}
E_{(x \sim U_{N})}[L_{C}(x)] \geq N
\end{equation}

In words the theorem states that
	no lossless compression program can achieve average codelengths smaller than $N$ bits,
	when averaged over all possible $N$ bit input strings.
In the subsequent development this is referred to as the ``No Free Lunch'' (NFL) theorem of data compression,
	as it implies that one can achieve compression for some strings $x$
	only at the price of inflating other strings.
At first glance, 
	this theorem appears to turn the CRM proposal into nonsense.
In fact, the theorem is the keystone of the CRM philosophy
	because it shows how lossless, large-scale compression research 
	must incorporate empirical science at a fundamental level.
	
To see this point, consider the following apparent paradox.
In spite of the NFL theorem, 
	lossless image compression programs exist and have been in widespread use for years.
As an example, the well-known Portable Network Graphics (PNG) compression algorithm
	seems to reliably produce encoded files that are 
	40-50\% shorter than would be achieved by a uniform encoding.
But this apparent success seems to violate the No Free Lunch theorem. 	

The paradox is resolved by noticing that the images used to evaluate
	image compression algorithms are not drawn from a uniform
	distribution $U_{N}(x)$ over images.
If lossless image formats were evaluated based on 
	their ability to compress uniformly random images, 
	no such format could ever be judged successful.
Instead, the images used in the evaluation process
	(which may be based on formal benchmarks or simple popularity)
	belong to a very special subset of all possible images:
	those that arise as a result of everyday human photography.
This ``real world'' image subset, though vast in absolute terms,
	is miniscule compared to the space of all possible images.
The NFL theorem permits compression for some special subset of inputs
	but requires in exchange the inflation of the non-special subset. 

In the light of the above remarks two equally valid perspectives
	on the image compression problem can be articulated.
In the first view the task of the compression researcher is to
	precisely define the special image subset to which the compressor will
	assign short codes at the expense of the non-special images.
For obvious technical reasons relating to 
	both the complexity and the feasibility of implementation of the compressor,
	this privileged subset must be defined in a minimalist or implicit manner
	and not by raw enumeration.
The subset specification must be done by listing properties or features that are
	present in the prefered images and absent in the non-prefered ones.
The goal then is to obtain an increasingly sophisticated understanding 
	of the characteristics of real world images
	by iterative refinement of the prefered image subset. 
Thus in this view research proceeds by proposing a precise and computationally tractable 
	definition of a prefered image subset and measuring,
	by means of the compression rate achieved on some benchmark database,
	the extent to which it overlaps with the real world or empirical image subset.

\begin{figure*}[t]
\centering
\subfigure{\includegraphics[width=.4\textwidth]{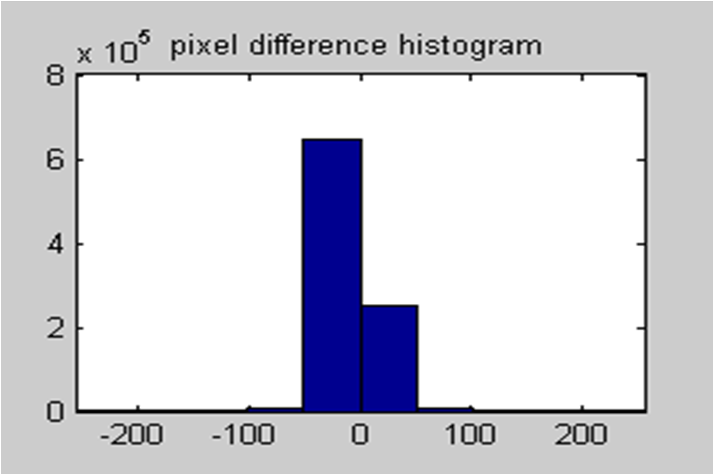}}
\subfigure{\includegraphics[width=.4\textwidth]{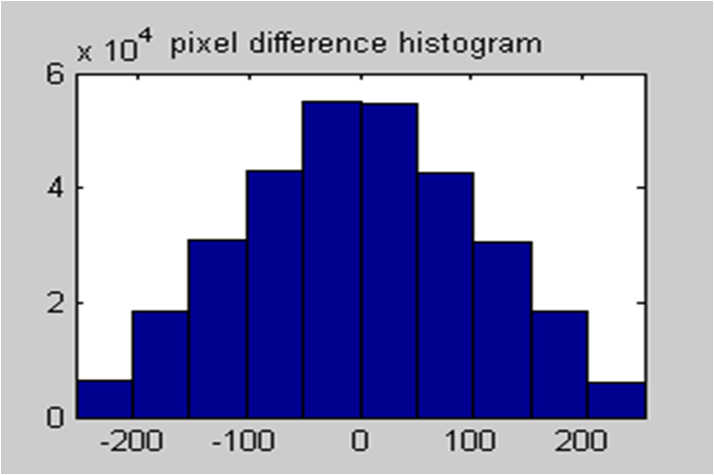}}
\caption[Pixel difference histograms for a natural and a random image.]{
Histograms of differences between values of neighboring pixels
	in a natural image (left) and a random image (right).
The clustering of the pixel difference values around 0 in the natural image
	is what allows compression formats like PNG to achieve compression.
Note the larger scale of the image on the left; 
	both histograms represent the same number of pixels.
}
\label{fig:pixeldiffn}
\end{figure*}

In the second perspective the task of the compression researcher
	is to discover structures or patterns in the real world images,
	and develop compressors capable of exploiting those features.
One obvious property of real world images is that
	adjacent pixels tend to have very similar values.
This property can be exploited by encoding the differences between neighboring
	pixel values instead of the values themselves.
The distribution of differences is very narrowly clustered around zero,
	so they can be encoded using shorter average codes.
Of course, this trick does not work for random images,
	in which there is no correlation between adjacent pixels
	(see Figure~\ref{fig:pixeldiffn}).
In this perspective research proceeds by discovering properties of 
	real world images and demonstrating how those properties can be exploited
	to achieve superior compression rates.

Both of the conceptualizations mentioned above depend strongly on 
	mathematical or computational techniques,
	either for the parsimonious specification of a prefered image subset
	or for the development of an algorithm that can exploit a certain type of image property.
But it is also clear that mathematical logic is only one of the two necessary ingredients,
	and in some sense the secondary or auxilliary component.
The primary component is the \textit{empirical} investigation
	required to discover the properties of real world images
	or to characterize the image subset to which they belong.
The data compression procedure recapitulates the procedure of physics:
	it begins with a hypothesis, 
	then develops a mathematical characterization of that hypothesis, 
	and finally proves the hypothesis by showing a correspondence between the
	mathematical theory and empirical reality.
Crucially, the hypothesis is not obtained through formal analysis
	or derivation and its sole ultimate justification is the 
	empirical correspondence shown in the final step.
	
\subsubsection{Falsifiability}

Throughout the history of science,
	philosophers have struggled to define what exactly science is 
	and how scientific theories can be separated from
	pseudoscientific ones. 
This is known as the problem of demarcation:	
	given a theory of astronomy and a theory of astrology,
	what justification can be used to identify the former as legitimate science
	and the latter as mere superstition?
Perhaps the most famous answer to this question was given by Karl Popper,
	who proposed to accept a theory as scientific
	if it was \textit{falsifiable}~\cite{Popper:1959}.
A falsifiable theory is one that makes explicit anti-predictions
	regarding the outcome that should be expected 
	for some experimental configuration. 
If the anti-predicted outcome actually occurs when the experiment is performed, 
	then the theory is discarded. 
A scientific field that traffics only in falsifiable theories 
	is able to conduct an efficient search through theory-space, 
	rapidly generating new candidates and discarding those 
	that do not agree with experimental observations~\cite{Platt:1964}.
	
The principle of demarcation provided by the CRM
	can be viewed as a continuous or graduated version of the Popperian principle,
	and comes about as a result of the intrinsic difficulty of lossless data compression
	in light of the No Free Lunch theorem.
Random data cannot be compressed.
In order to achieve compression on a real data set,
	it is necessary to reassign probability away from certain outcomes
	and toward other outcomes.
This probability reassignment requirement is just a 
	softened version of the Popperian anti-prediction requirement. 
Just as Popper designates as non-scientific theories that do not make anti-predictions, 
	the CRM designates as non-scientific theories that do not make probability reassignments,
	since such theories cannot achieve compression.
Any proponent of a new scientific theory justified by the CRM
	runs the risk of embarassment if it turns out that the theory 
	not only fails to compress the dataset but actually \textit{inflates} it.

The crucial difference between Popperian science and CRM science
	is that the former appears to justify stark binary assessments
	regarding the truth or falsehood of a theory,
	while the latter provides only a number which can be compared to other numbers.
If theories are either true or false, 
	then the compression principle is no more useful than the falsifiability principle.
But if theories can exist on some middle ground between absolute truth and its opposite,
	then it makes sense to claim that one theory is relatively more true than another,
	even if both are imperfect.
The compression principle can be used to justify such claims.
Falsifiability consigns all imperfect theories to the same garbage bin;
	compression can be used to rescue the valuable theories from the bin,
	dust them off, and establish them as legitimate science.

Another difference between the falsifiability principle and the compression principle
	is that the former appears to allow a theory to be evaluated
	in absolute terms, without reference to any other theory.
The CRM assigns a score to an individual theory, 
	but the score is only useful for the purposes of comparison, 
	and provides very little insight into the absolute quality of the theory. 
In practice, though, 
	the actual function of empirical validation is to select
	between rival theories. 	
In real situations that require decision-making, 
	it is necessary to reify some theory to the status of champion,
	even if the ``theory'' is merely a formalized admission of total ignorance
	regarding the phenomenon.
In the Popperian view, 
	the champion theory is the one that has withstood all attempts at falsification.
In the CRM view, 
	the champion theory is the one that achieves the smallest codelength
	on the relevant benchmark database.

\section{Compression and Computer Vision}
\label{sec:comprvsion}

The goal of computer vision research is to build machines 
	that can see as well as humans.
For many who are not familiar with the field, 
	this seems like a rather simple goal. 
But it is in fact profoundly difficult, 
	and in spite of several decades of research,
	current vision systems cannot perform
	even at the level of human children.
One might diagnose a number of factors
	contributing to this lack of progress.
It could be that the mathematical theory employed by the field is not 
	sophisticated enough, or that modern computers are not fast enough. 
But the present paper suggests an alternative hypothesis,
	which is that the problem lies in the philosophical foundation of the field.

The words ``philsophical foundation'' 
	are used here to refer to the answers to two very concrete questions.
The first is the question of evaluation:
	given a pool of candidate solutions to a certain task,	
	how does one select the best one?
As discussed in Section~\ref{sec:comprvsion},
	there is a substantial amount of work in the area of empirical evaluation,
	but it is safe to say that current methods are not completely satisfying,
	either from a conceptual or a practical standpoint.
The philosophical difficulties involved in the evaluation of computer vision solutions
	stands in stark contrast to the conceptual ease with which 
	two theories of physics can be compared,
	by finding an experimental configuration for which the two theories 
	make conflicting predictions and running the experiment.

Behind the question of evaluation lies
	another, deeper question.
This is the meta-question: 	
	what are the important problems in the field, and why are they important?
Computer vision
	currently can provide only a litany of relatively weak answers to this question.
This paper proposes a decisive, ``purist'' answer to the meta-question:	
	the goal of vision is to describe visual reality,
	and a problem is important if a solution to it can be used
	to achieve an improved compression rate on a vast database of natural images.
Section~\ref{sec:advantages} discusses several advantages
	achieved by formulating vision research in this way.
A crucial argument, developed in Section~\ref{sec:equivalence},
	is that traditional computer vision research is deeply related 
	to image compression.
Many standard vision tasks can be refomulated
	as specialized image compression techniques.
Perhaps the most important point, however,
	is that the new view allows computer vision to be formulated
	as a hard empirical science, like physics or chemistry,
	in which theories are tested, refined and often discarded.
A segmentation algorithm makes no quantifiable assertion about empirical reality,
	but a segmentation-based compressor does.
	
\subsection{Shortcomings in Current Evaluation Methodology}
\label{sec:mthoddiffc}

Evaluation is one of the central conceptual difficulties in computer vision;
	this section presents a brief analysis and critique of current methods.
Since each computer vision task requires its own evaluation procedure (or ``evaluator'')
	to judge candidate solutions, and some have more than one,
	there are at least as many evaluators as there are tasks.
A comprehensive survey is therefore beyond the scope of the paper,
	which instead provides a discussion that should 
	illustrate the basic issues.
Before proceeding to the discussion, however,
	it is worth noting that, historically at least,
	the field has exhibited very bad habits 
	regarding the issue of empirical evaluation.
Shin \etal~\cite{Shin:1998} note that, 
	of 23 edge detection papers that were published in four journals between 1992 and 1998, 
	not a single one gave results using ground truth data from real images.
Even when empirical evaluations are carried out,
	it is often by a research group that has developed
	a new technique and is interested in highlighting its performance
	relative to existing methods.
The inadequacies of the evaluation methodology currently employed in computer vision
	have been lamented at length by several authors~\cite{Haralick:1986,Jain:1991}.
	
To begin the discussion, consider the task of image segmentation,
	which illustrates several of the challenging issues 
	that hamper evaluation work in the field as a whole.
Segmentation research dates back at least to 1978~\cite{Ohlander:1978}.
For many years, solutions were evaluated primarily by showing
	the outputs of an algorithm when applied to a small number of images,
	and appealing to the reader to confirm that the results
	agreed with human perception.
A slightly more rigorous evaluation methodology
	involved numerical scores based on various aspects of the segmentation
	(see the review by Zhang~\cite{Zhang:1996}).
But the use of such scores is somewhat circular:
	one can easily define an algorithm that works by optimizing the score,
	and it would then presumably do better than any other technique.
	
Finally, in 2001, a real empirical benchmark appeared in the form 
	of the Berkeley Segmentation Dataset~\cite{Martin:2001},
	which contains ``ground truth'' obtained by asking human test subjects
	to segment images by hand.
While the approach to evaluation associated with this dataset
	is an important improvement over previous techniques, several issues remain.
One obvious problem with this approach is that it is 
	highly labor-intensive and task-specific,
	so the ratio of effort expended to understanding achieved seems low.
A larger issue is that the segmentation problem has no precisely defined correct answer:
	different humans will produce substantially different responses 
	to the same image.
Even this might not be so bad;
	one can define an aggregate or average score and 
	plausibly hope that using enough data will damp out chance fluctuations
	that might cause a low quality algorithm to achieve a high score or vice versa.
But still another conceptual hurdle must be cleared:
	given two segmentations, one algorithm- and one human-generated,
	there is no standard way to score the former by comparing it to the latter.
Some scoring functions assign high values to degenerate responses, 
	such as assigning each pixel to its own region,
	or assigning the entire image to a single region~\cite{Martin:2002}.
The question of how to score a segmentation by comparing it to 	
	a human-produced result has become a research problem 
	in its own right, 
	resulting in a proliferation of scoring methods~\cite{Martin:2002,Estrada:2005,Unnikrishnan:2007}.
A more technical but still important issue is the problem of parameter choice.
Essentially all segmentation algorithms require 
	the choice of at least one, and usually several, parameters,
	which strongly influence the algorithm's output.
This complicates the evaluation process for obvious reasons.

The task of edge detection is conceptually similar to segmentation,
	and faces many of the same issues when it comes to empirical evaluation.
As with segmentation, there is ambiguity in the problem definition:
	is the goal to recover perceptually salient edges?
If not, then what criterion can be used to determine if, say,
	the Canny edge detector is better than the Sobel operator?
If so, then one can in principle evaluate detectors by 
	comparing their output to human-generated edge responses.
This is the approach taken by Bowyer~\etal~\cite{Bowyer:1999}.
A major drawback of this approach is the amount of human effort required:
	the authors report that it takes 3-4 hours to provide
	ground truth for a single 512x512 image.
This inevitably implies that only a small number of images can be used in the evaluation;
	~\cite{Bowyer:1999} uses 40 images,
	from four different image categories.
The issue of parameter choice is especially acute in this domain.
The authors of~\cite{Bowyer:1999} use a complicated parameter sampling scheme 
	based on receiver operating characteristic (ROC) curves
	to find good settings for a given image.
However, as shown by Forbes and Draper~\cite{Forbes:2000},
	at least in the case of the Canny detector,
	small changes in the input image can result in large
	changes in the edge detector response.
This suggests that the conclusion of~\cite{Bowyer:1999},
	that the Canny achieved the best results,
	could in fact be due to the combination of the adaptive parameter sampling scheme
	with the Canny's large sensitivity to parameter values.
Bowyer~\etal~\cite{Bowyer:1999} claim that the performance results
	were mostly consistent over several image categories, 
	but this could be a statistical fluke due to the small
	number of images used.
This consistency result also contrasts strongly with the result of Heath~\etal~\cite{Heath:1996},
	which found that for several categories,
	there were no statistically significant differences in the 
	performance of the edge detectors.

A third standard task in computer vision is object recognition,
	which seems to support a straightforward evaluation procedure:
	construct a labeled database
	and measure the number of errors made by each candidate solution.
In spite of this seeming simplicity,	
	evaluation of object recognition systems is a quite challenging problem~\cite{Ponce:2006}.
One issue relates to the distinction between inter-class and intra-class variation.
As an example, the popular Caltech101 database~\cite{Feifei:2007}
	contains a broad range of object categories 
	(including ``brontosaurus'', ``euphonium'', and ``Garfield''),
	so that inter-class variation is large.
However, the images show mostly just the object of interest,
	centered and without much clutter or occlusion,
	thus minimizing intra-class variation.
Since inter-class variation is large while intra-class variation is small,
	the objects are much easier to recognize than they would be in a real world application.
Another issue is parameter overfitting:
	the publication of a benchmark like Caltech101 generally sparks a flurry of new papers,
	which report progressively better results over a several year span.
This improvement might not be due to better technology,
	but instead to the fine-tuning of algorithm parameters. 
Reports of exceptional performance are also somewhat misleading:
	given that some techniques achieve 95\% accuracy on some databases~\cite{Ponce:2006},
	an uninitiated observer might conclude that the problem of object recognition 
	has been solved, which is obviously not true.
But if excellent benchmark performance does not imply excellent real-world performance,
	what exactly has been learned by using the benchmark?
Related to this discrepancy is the fact that several techniques can
	exhibit excellent classification performance but very poor localization performance
	(they can guess that the image contains a chair, but don't know where the chair is).
This is partly because, in some benchmark databases,
	the identity of the target object correlates highly with the background.
The correlation improves the performance of ``global'' methods 
	that use information extracted from the entire image.
To some extent this makes sense, 
	because for example cows are often found in grassy pastures,
	but a recognizer that exploits this information may fail to 
	recognize a cow in a photograph of an Indian city.
	
A final area of consideration involves the two conceptually similar
	tasks of stereo matching and optical flow estimation.
The key property of these tasks is that there exists an objectively correct answer,
	and ground truth corresponding to this answer can be obtained.
The evaluation methods 
	proposed in the literature for these tasks can be broken down into two types
	\cite{Scharstein:2002,Szeliski:1999,Scharstein:2003,Baker:2007}.
The first type of evaluator depends on using a sophisticated experimental
	apparatus to obtain the ground truth.
For the stereo matching problem,
	ground truth can be obtained using an apparatus that employs structured light~\cite{Scharstein:2003}.
For the optical flow problem,
	an experimental setup is used in which an object sprinkled with flourescent
	paint is moved on a computer-controlled motion stage,
	while being photographed in both ambient and ultraviolet lighting~\cite{Baker:2007}.
Once the ground truth has been obtained,
	it is a conceptually simple matter to evaluate a solution
	by comparing its output to the correct answer.
The major drawback to this approach 
	is the difficulty of using the experimental apparatus,
	which implies that only a small number of image sequences are used.
A well-known benchmark, hosted on the Middlebury Stereo Vision Page,
	contains a total of 38 sequences~\cite{Scharstein:2010}.
Given that most vision techniques involve multiple parameters,
	it seems hard to rule out overfitting as the source of 
	any good performances achieved on such a small dataset.
	
For both of the tasks mentioned above, however,
	there exists a strikingly simpler evaluation 
	method~\cite{Scharstein:2002,Szeliski:1999,Baker:2007}.
The basic idea is to use the output of the candidate solution to infer or estimate
	a new unseen image, and compare the predicted image to the real thing.
The paper~\cite{Baker:2007} proposes the following evaluation metric
	for the optical flow task.
First an image sequence is grabbed at 100 Hz.
Then a subsequence corresponding to every fourth image
	(implying a frame rate of 25 Hz) 
	is fed to the optical flow algorithm.
Based on the resulting flow estimates,
	interpolation is used to predict the unseen images in the 100 Hz sequence,
	and a prediction error is computed.
A similar interpolation-scheme evaluator for the stereo matching task 
	is proposed in~\cite{Scharstein:2002,Szeliski:1999}.
Several image sets are obtained using a trinocular camera.
Then, using the extremal images as inputs to the stereo matching algorithm,
	its output is used to infer the middle image,
	which is used to calculate the prediction error.
One simple function for the prediction error is the RMSE;
	another is a modified version of the RMSE
	that depends also on the local image gradient.
These evaluator methods have a decisive advantage in that they 
	do not require any special equipment,
	and so in principle large quantities of data can be used.
Furthermore, as discussed in Section~\ref{sec:equivalence},
	these scores correspond almost exactly to codelengths
	that would be achieved by a certain type of compression algorithm.
	
\subsection{Concrete Proposal: Highway Camera}
\label{sec:cncrtprpsl}

This section turns the abstract idea of applying the CRM to large databases of natural images
	into a concrete proposal, 
	by supplying specific details related to one way of constructing a database
	and using sophisticated computational tools to compress it.
The proposal calls for setting up a video camera next to a highway,
	and producing a target database by taking video from the stream of passing cars.
Since the camera does not move, 
	and there is usually not much activity on the sides of highways, 
	the main source of variation in the resulting video will be the automobiles.
Therefore, in order to compress the video stream well, 
	it will be necessary to obtain a good computational understanding 
	of the appearance of automobiles.
	
A simple first step would be to take advantage of the fact that cars
	are rigid bodies subject to Newtonian laws of physics.
The velocity of a car must be a continuous function of time.
Given a series of images at timesteps $\{t_{0}, t_{1}, t_{2} \ldots t_{n}\}$
	it is possible to predict the image at timestep $t_{n+1}$
	simply by isolating the moving pixels in the series 
	(these correspond to the car), 
	and interpolating those pixels forward into the new image,
	using basic rules of camera geometry and calculus.
Since neither the background nor the moving pixel blob changes much between frames, 
	it should be possible to achieve a good compression
	rate using this simple trick.

Further improvements can be achieved by detecting and exploiting patterns
	in the blob of moving pixels.
One observation is that the wheels of a moving car have
	a simple characteristic appearance - 
	a dark outer ring corresponding to the tire,
	along with the off-white circle of the hubcap at the center.
Because of this characteristic appearance,
	it should be straightforward to build a wheel detector
	using standard techniques of supervised learning.
One could then save bits by representing the wheel pixels
	using a specialized model.
Further progress could be achieved by conducting 
	a study of the characteristic appearance of the surfaces of cars.
Since most cars are painted in a single color,
	it should be possible to develop a specialized algorithm
	to identify the frame of the car.
Extra attention would be required to handle the complex reflective appearance
	of the windshield.
Note that the encoder always has the option of ``backing off'';
	if attempts to apply more aggressive encoding methods fail
	(e.g., if the car is painted in multiple colors),
	then the simpler pixel-blob encoding method can be used instead.
	
Additional progress could be achieved by recognizing that
	most automobiles can be categorized 
	into a discrete set of categories (e.g., a 2009 Toyota Corolla).
Since these categories have standardized dimensions,
	bits could be saved by encoding the category of a car
	instead of information related to its shape.
Initially, the process of building category-specific modules 
	for the appearance of a car might be difficult and time-consuming.
But once one has developed modules for the Hyundai Sonata, Chevrolet Equinox,
	Honda Civic, and Nissan Altima,
	it should not require much additional work to construct
	a module for the Toyota Sienna.
Indeed, it may be possible to develop a learning algorithm that,
	through some sort of clustering process,
	would automatically extract,
	from large quantities of roadside video data,
	appearance modules for the various car categories.
	
The above discussion illustrates several important aspects
	of the proposed inquiry.
A crucial difference between the traditional approach and the CRM approach 
	to vision research is the shift in mindset away from pure theorizing
	and towards the study of specific structures in visual reality.
Empirical details such as hubcaps or windshields almost
	never take center stage in traditional computer vision papers - 
	they are either buried in the results section or ignored entirely.
The key insight is that theories of these details can be developed
	and rigorously evaluated.
On the other hand, the new approach \textit{does not} dismiss the possibility 
	of abstract, automated learning methods.
These methods are in some sense the real goal,
	but their development is postponed until more basic understanding can be achieved.
It should also be clear that, if successful, the type of inquiry outlined
	above will yield practical results,
	in the form of sophisticated vision systems that can be used in robotic cars.

The concrete proposal is useful for another reason:
	it allows skeptical observers to formulate their objections in precise terms.
One such objection attacks a key implicit hypothesis of the CRM,
	which is that improved abstract understanding will produce improved compression rates.
In other words, 
	in order to achieve the best possible codelengths,
	it will be necessary to employ abstractions such 
	as the position, velocity, outline, and model of a car.
If this hypothesis is incorrect,
	then the fruit of CRM research will be of very dubious value.
Another potential objection is simply that the reasoning behind the proposal
	underestimates the real difficulty of the problem,
	thus committing the same error that was made in the early days of the field.
It is possible that even 
	the simplified roadside video data will contain 
	so much variation that it will be impossible to make progress.
Or, similarly, that the inquiry will basically work,
	but will require far more effort than is justified by
	the results it could potentially achieve.
These objections are plausible but not invincible,
	and do not disallow a reasonable hope of success.
	
\subsection{Advantages of Vision-as-Compression Approach} 
\label{sec:advantages}

The proposal of this paper is to approach vision science
	by applying the Compression Rate Method to large databases of natural images.
The present section lists several advantages of this idea.
	
\subsubsection{Methodological Efficiency}

One can imagine an academic journal dedicated to CRM research that accepts
	two types of papers.
The first type includes reports of new shared image databases for use by the community.
These submissions are accompanied with the actual database,
	which the journal editors briefly inspect,
	and then publish on the journal web site 
	where it can be downloaded by other interested researchers.
The second type of submission includes reports of new compression rates
	achieved on one of the shared databases.
A submission of this type must be accompanied by the actual compressor
	used to achieve the reported result.
As part of the review process, 
	the journal editors run the compressor, 
	verify that the real net codelength agrees with the reported result,
	and then check that the decoded version matches the original.
This ``trust, but verify'' approach should provide a strong degree of
	protection against both honest mistakes and actual academic fraud.
In principle, 
	the editors of this journal have an easy job,
	since any paper that legitimately reports an improved compression result should be accepted.
In practice, it may be necessary for the editors to exercise some degree of 
	qualitative judgment: 
	a paper that merely tweaks some settings of a previous result
	and thereby achieves a small reduction in codelength is probably not worth publishing,
	while an innovative new approach that doesn't quite manage to
	surpass the current champion probably is 
	(new techniques will probably require some polishing and refinement 
	before they become competitive).
The ease of evaluating new research is one of the central advantages of the CRM
	approach to vision.
		
A related advantage is the relative ease of building large datasets.
A major problem for traditional computer vision is 
	the effort required to build ground truth databases.
The CRM mandates the use of unlabeled databases,
	which are much easier to construct.
Of course, it will be necessary to exercise some ingenuity and foresight
	when building target databases, especially in the early stages of research.
It is probably impossible to make a lot of progress
	simply by using, say, the database of images
	hosted on Flickr.com.
The degree of variation in such an image collection 
	seems unapproachably large.
Instead, it will probably be more fruitful to 
	use a simplified database that exhibits a smaller degree of variation; 
	see Section~\ref{sec:cncrtprpsl} for one proposal.
	
\subsubsection{Scalable Evaluation Paradigm} 

The word scalability refers to the rate at which 
	the cost of a system increases as increasing demands are placed on it.
For example, 
	a process in which widgets are built by hand is not very scalable, 
	since doubling the number of widgets will approximately double 
	the amount of labor required.
A widget factory, in contrast,
	might be able to double its output using only a ten percent increase in labor, 
	and is therefore highly scalable.
The argument of this section is that the current evaluation paradigm of computer vision
	has very bad scaling properties,
	while the CRM approach to evaluation is highly scalable.
	
The current evaluation paradigm can be thought of as ``one-to-one'':
	each specific vision task, 
	such as object recognition, image segmentation, or stereo matching, 
	requires its own evaluator method.
This one-to-one methodology scales badly,
	since the process of developing an evaluator is difficult work,
	requiring both intellectual, and often manual, labor.
Furthermore, this arduous labor may very well go unrewarded.
Sometimes, after substantial effort has been invested in developing an evaluator,
	it turns out the scheme suffers from some flaw
	(a possible example of this is the difficulty related
	to the ROC curve evaluation scheme~\cite{Bowyer:1999,Forbes:2000}).
Also, it is not obvious that modern evaluators 
	assign consistently higher scores to higher-quality solutions.
It may be necessary 
	to conduct a meta-evaluation process
	in order to rate the quality of the evaluators.
Because developing an evaluator is risky and difficult,
	the current one-to-one evalustor paradigm is painfully inefficient.	

The importance of scalability becomes even more obvious 
	when one realizes that many vision tasks of current interest,
	such as image segmentation, edge detection, and optical flow estimation,
	are low-level, and not directly useful.
Instead, the idea is that once good solutions are obtained for the low-level tasks, 
	they can be incorporated into higher-level systems.
These more advanced systems, then, are the real ultimate goal of research. 
Of course, under the current paradigm,
	each high-level task will require its own evaluator method
	to compare candidate solutions.	
So the mountain of work required to complete the project of 
	finding evaluators for \textit{current} tasks
	is tiny compared to the mountain of work 
	that will be required to develop evaluators for \textit{future} tasks.
	
In contrast to the current paradigm,
	the Compression Rate Method provides the ability to evaluate a large
	number of disparate techniques using a single principle.
Visual reality contains an practically unlimited number of empirical regularities,
	each of which can be characterized and exploited to save bits.
As discussed in the proposal of Section~\ref{sec:cncrtprpsl},
	a single database of roadside video can be used to 
	evaluate the performance of a large number of components,
	such as motion detectors, wheel detectors, specialized segmentation algorithms,
	and learning algorithms that infer car categories.

In addition to greatly reducing the number of man-hours required for evaluator development, 
	the compression principle provides a decisive answer 
	to the question of what the high-level tasks \textit{are}.
According to the CRM,
	a low-level system is one that achieves compression by exploiting
	simple and relatively obvious regularities,
	such as the fact that cars are rigid bodies obeying Newtonian laws of motion.
A higher-level system is built on top of lower-level systems
	and achieves an \textit{improved} compression rate 
	by exploiting more sophisticated abstractions,
	such as the fact that cars can be categorized by make and model.

\subsubsection{Justification of Complex Models}

\begin{figure}
\begin{centering}
\includegraphics{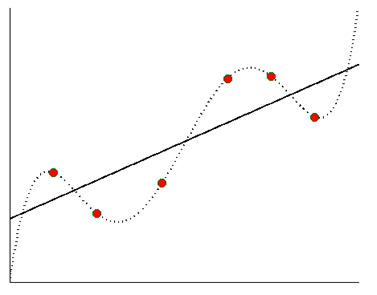}
\caption{
Illustrating the complex model penalty idea:
	in this low-data regime the simple line model should be prefered,
	even though the complex model achieves zero error.
}
\label{fig:qrticpolyn}
\end{centering}
\end{figure}

A major concern in the field of machine learning
	is a phenomenon called \textit{overfitting}.
Since many computer vision applications rely on machine learning methods,
	this problem is relevant to computer vision as well.
Overfitting is said to occur when a model achieves very good performance
 	on the known (``training'') data,
	but fails to generalize to unseen (``test'') data.
The basic cause of overfitting is that the model is too complex
	relative to the data it describes.
This situation is illustrated in Figure~\ref{fig:qrticpolyn}.

One of the great achievements of machine learning is the discovery 
	of methods to prevent overfitting.
These methods are all similar in spirit:
	the secret is to apply a penalty to complex models,
	which is more or less severe depending on the amount of data.
The technical issue is how to quantify the complexity of a model
	in relation to the volume of data.
In the VC-theory~\cite{Vapnik:1998},
	a quantity called the VC-dimension
	is used to characterize model complexity
	and derive generalization bounds.
One typical VC-theory generalization bound is the following inequality:

\begin{equation}
R(g) \leq R_{n}(g) + 2 \bigg ( 2 \frac{h \log \frac{2en}{h} + \log \frac{2}{\delta}}{n} \bigg)^{\frac{1}{2}}
\end{equation}

Where $R(g)$ is the real performance of a model $g$,
	$R_{n}(g)$ is the empirical performance,
	$n$ is the number of training samples,
	$h$ is the VC-dimension of the model class,
	and the inequality holds with probability $1 - \delta$.
Thus, if the model complexity term $h$ is small compared
	to the number of samples $n$,
	the bound is tight and the real performance should be 
	about the same as the empirical performance
	(technically $R$ stands for risk, and the goal is to minimize risk,
	so the fear is that $R(g)$ could be much larger than $R_{n}(g)$).

Another formulation of the model complexity idea,
	which as mentioned above is highly related to the CRM, 
	is the Minimum Description Length (MDL) principle~\cite{Rissanen:1978}.
Here the goal is to idea is to compress a data set to the smallest possible size.
If a good model can be found,
	it can be used to encode the data with a short code.
But one must also account for the number of bits required to encode the model itself.
The result is a tradeoff between model complexity and empirical performance,
	just as in the VC generalization inequality shown above.
	
A key point is that both of these ideas specifically \textit{allow}
	the use of complex models.
The only requirement is that a complex model be justified 
	by a correspondingly large amount of data.
In terms of Figure~\ref{fig:qrticpolyn},
	if there were thousands of data points and they all fell on the quartic polynomial, 
	one would certainly be justified in using it as a model.

Computer vision applications that are built on top of machine learning algorithms
	must follow the same basic rules regarding model complexity,
	or suffer from overfitting.
The crucial question is: 
	what is the relevant data set against which model complexity
	must be justified?
For most computer vision systems,
	the answer is the labeled database used to train the classifier.
Since labeled databases are usually quite small,
	this implies that the resulting models must be very simple.

However, a critical but underappreciated characteristic of 	
	computer vision is that it is \textit{easy} to obtain 
	large quantities of raw data.
Only labeled data is in short supply.
In the CRM, complex models can be justified by showing
	that they achieve net codelength reductions
	when applied to large databases of raw, unlabeled data.
For example, one could justify the use of a model requiring 100 Mb to specify
	(larger by orders of magnitude than the models used in typical learning research)
	by showing that it saves 500 Mb when used to encode a 10 Gb image database.
Of course, simpler is always better: 
	if a 10 Mb model can be used to achieve the same savings,
	it should be preferred.
	
Similar articulations of the idea that more complex models can 
	be justified when modeling raw data have appeared in the literature. 
For example, Hinton \etal~note that:

\begin{quote}
Generative models can learn low-level features without requiring feedback from the label, 
	and they can learn many more parameters than discriminative models without overfitting. 
In discriminative learning, 
	each training case constrains the parameters only by as many bits of information 
	as are required to specify the label. 
For a generative model, each training case constrains the parameters by the 
	number of bits required to specify the input~\cite{Hinton:2006b}.
\end{quote}

The distinction between generative models and discriminative models
	is roughly similar to the distinction between unsupervised CRM-style learning
	and traditional supervised learning.
Hinton develops the generative model philosophy at 
	greater length in~\cite{Hinton:2007}.
Note that a generative model is exactly one half of a compression program 
	(the decoder component).

\subsubsection{Systematic Progress}

Thomas Kuhn, in his essay on the history of science, posed a famous rhetorical question: 
	``Is a field a science because it makes progress, or is it a science because it makes progress?''
	\cite{Kuhn:1970}.
In other words, 
	do true scientific disciplines possess some special methodological characteristic
	that allows them to make progress,
	or is science simply defined as the specific subset of academic research
	that shows continual improvement?
Kuhn's point in asking this question was not to find an answer,
	but merely to illustrate that almost any field that 
	actually makes cumulative progress is also a science, and vice versa.
Computer vision may make progress -
	certainly new and impressive applications appear at a fairly steady rate.
But it is not clear if it makes \textit{systematic} progress.
		
One sign that the field of computer vision has a problem in this regard
	is the constant replication of effort.
For any given task,
 	such as segmentation, edge detection, annotation, or image reconstruction,
	there are dozens, if not hundreds, of candidate solutions 
	to be found in the literature.
In and of itself, that may not be so bad;
	any unexpected new experimental observation in physics might 
	merit a large number of proposed theoretical explanations.
The difference is that the physicists are able to ultimately 
	find the \textit{right} explanation,
	at which point people can go on to new areas of research.
	
Systematic progress is built into the very definition of the 
	Compression Rate Method.
In effect, an investigation guided by the compression principle	
	\textit{must} make clear, quantifiable progress, or grind to a complete halt. 
By justifying strong comparisons between competing theories,
	the CRM enables the community to conduct a rapid search through the space of theories.
The ability to search systematically through the theory-space is 
	one of the reasons fields like physics can make rapid progress~\cite{Platt:1964}.
	
The method also provides an easy mechanism for researchers 
	to build on each others' work.
Given an algorithm that achieves a certain level of performance,
	the new researcher can simply add a specialized module
	that works for certain types of images, and lies dormant for others.
As an example, a researcher could take a generic image compressor such as PNG,
	and add a special module for encoding faces (see Section~\ref{sec:facedetmod}).
The enhanced compressor would then achieve must better codelengths 
	for images that include faces.
Another researcher could then add a module for encoding arms and hands
	(the pixels in these regions would be narrowly distributed around some mean skin color).
Of course, the improvement achieved by any single contribution may not be huge,
	but this mode of research is guaranteed to achieve cumulative progress.
	
In addition to its ability to measure and facilitate progress,
	the compression idea provides another subtle advantage.
It provides a software engineering principle to
	guide the integration of a large number of separate computational modules.
Given a set of software modules providing traditional implementations of 
	segmentation, object recognition, edge detection, camera calibration, and so on,
	it is not at all obvious how to package these modules 
	together into one integrated system.
An attempt to do so would likely result in little more than a package of libraries,
	rather than a real application with a clear purpose.
Establishing compression as the function of the application
	provides a principle for binding the modules together.
The input is the raw image, 
	the output is the encoded version,
	and the software modules are employed in various ways
	to facilitate the transformation.

Is the ability to make systematic progress \textit{sufficient} 
	for a field to be considered a science?
The answer might depend on personal definitions.
One might hesitate to call the subfield of AI chess a science,
	if for no other reason than that it involves a purely artifical game
	that has no connection to the real world.
But chess research makes systematic progress,
	since the game of chess supports strong comparisons of rival solutions.
It is not a coincidence that this subfield 
	has also produced one of the most dramatic demonstrations
	of the potential of artificial intelligence.
The approach to vision science proposed in this paper
	supports systematic progress while also 
	mandating a comprehensive interrogation of an important aspect of physical reality.
	
\subsection{Reformulation of Vision as Compression Problem}
\label{sec:equivalence}

\subsubsection{Abstract Framework}

Computer vision is often described as the inverse problem of computer graphics.
The typical problem of graphics is to produce,
	given a scene description $D_{L}$ wrtten in some description language $L$,
	the image $I$ that would be created if a photo were taken of that scene.
The goal of computer vision is to perform the reverse process:
	to obtain a scene description $D_{L}$ 
	from the raw information contained in the pixels of the image $I$.
This goal can be formalized mathematically 
	by writing $I = G(D_{L}) + I_{C}$
	where $G(D_{L})$ is the image constructed by the graphics program
	and $I_{C}$ is a correction image that makes up for any discrepancies.
Then the goal is to make the correction image as small as possible:

\begin{eqnarray*}
D_{L}^{*} &=& \arg \min_{D_{L}} C_{disc}(I_{c}) \\
 &=& \arg \min_{D_{L}} C_{disc}(I - G(D_{L}))
\end{eqnarray*}

Where $C_{disc}$ is some cost function which is minimized for the zero image,
	such as the sum of the squared values of each correction pixel.
The problem with this formulation 
	is that it ignores one the major difficulties of computer vision,
	which is that the inverse problem is underconstrained:
	there are many possible scene descriptions that can produce the same image.
So it is usually possible to trivially generate any target image 
	by constructing an arbitrarily complex description $D_{L}$.
As an example, if one of the primitives of the description language is a circle, 
	and the circle primitive has properties that give its color and position relative to the camera, 
	then it is possible to generate an arbitrary image by 
	positioning a tiny circle of the necessary color at each pixel location. 	
The standard remedy for the underconstrainment issue 
	is regularization~\cite{Poggio:1989}.
The idea here is to introduce a function $h(D_{L})$ 
	that penalizes complex descriptions.
Then one optimizes a tradeoff between descriptive 
	accuracy and complexity:
	
\begin{equation}
D_{L}^{*} = \arg \min_{D_{L}} C_{disc}(I_{C}) + \lambda h(D_{L})
\end{equation}
	
Where the regularization parameter $\lambda$ controls how 
	strongly complex descriptions are penalized.
While this formulation works well enough in some cases,
	it also raises several thorny questions related
	to how the two cost functions should be chosen.
If the goal of the process is to obtain descriptions that appear 
	visually ``correct'' to humans, 
	then presumably it is necessary to take considerations
	of human perception into account when choosing these functions.
At this point the typical approach is for the practitioner to choose
	the functions based on taste or intuition,
	and then show that they lead to qualitatively good results.

It turns out that the regularization procedure
	can be interpreted as a form of Bayesian inference.
The idea here is to view the image as evidence
	and the description as a hypothesis explaining the evidence.
Then the goal is to find the most probable hypothesis given the evidence:

\begin{eqnarray}
D_{L}^{*} &=& \arg \max_{D_{L}} p(D_{L}|I) \nonumber \\
&=& \arg \max_{D_{L}} p(I|D_{L}) p(D_{L}) \nonumber \\
&=& \arg \min_{D_{L}} -\log p(I|D_{L}) -\log p(D_{L}) \nonumber \\
&=& \arg \min_{D_{L}} C_{disc}(I_{C}) + h(D_{L}) \label{eq:regularized}
\end{eqnarray}

In words, by identifying the conditional probability of an image 
	given a description with the discrepancy cost function
	($-\log p(I|D_{L}) = C_{disc}(I_{C})$),
	and the prior probability of a description with the regularization function
	($-\log p(D_{L}) = h(D_{L})$),
	the regularized optimization procedure is transformed into a
	Bayesian inference problem.	
This arrangement has the benefit of eliminating the $\lambda$ parameter,
	but sheds no light on the problem of selecting the two crucial functions.

But more insight can be gained by analyzing the problem in terms of data compression.
Consider a sender and a receiver who have agreed to transmit images
	using an encoding scheme based on the graphics program and
	the associated description language $L$.
The sender first transmits a scene description $D_{L}$,
	which the receiver feeds to the graphics program to 
	construct the uncorrected image $G(D_{L})$.
The sender then transmits the correction image $I_{C}$,
	allowing the receiver to losslessly recover the original image.
The parties have agreed on a prior distribution $p(D_{L})$ for the scene descriptions, 
	and a method of encoding the correction image that requires 
	a codelength of $C_{enc}(I_{C})$.
The goal is to find a good $D_{L}^{*}$ that minimizes the total codelength:

\begin{equation}
\label{eq:mdl}
D_{L}^{*} = \arg \min_{D_{L}} \big( C_{enc}(I_{c}) -\log_{2} p(D_{L}) \big)
\end{equation}	

This formulation of the problem is thus equivalent to Equation~\ref{eq:regularized},
	showing that the general problem of computer vision can 
	be formulated in terms of compression.
If the procedure is only going to be applied to a single image,
	then this perspective is not much better than the previous one.
But if many images are going to be sent,
	then this formulation provides a clean principle for
	selecting the prior and the cost function:
	they should be chosen in such a way as to minimize the total cost
	for the entire database.

\begin{figure}
\begin{centering}
\includegraphics[width=.5\textwidth]{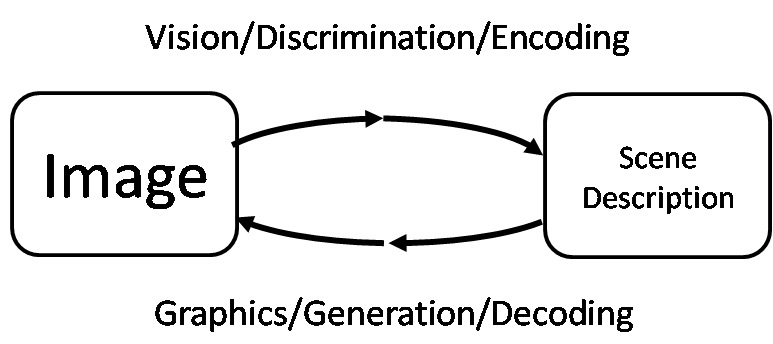} 
\caption{
Inverse relationship of graphics and vision.
}
\label{fig:backnforth}
\end{centering}
\end{figure}

This abstract analysis shows that there is 
	another, deeper problem in computer vision that is rarely addressed
	because the standard problem is hard enough.
This is the problem of choosing a description language $L$.
It is not obvious how the traditional conceptual framework
	of computer vision can be used to solve the problem of choosing $L$.
In contrast the CRM provides a direct answer:
	given two description languages $L_{a}$ and $L_{b}$,
	prefer the one that can be used to obtain better compression rates.
Note how this criterion simultaneously evaluates two computational tools: 
	the description language and the algorithmic methods used to 
	obtain actual descriptions.
Perhaps counterintuitively, 
	it may be the case that a simplistic description language
	may be selected by the compression principle over a
	more realistic, full-bodied language if the former supports
	a better inference algorithm.
	
The two formulations of the vision problem discussed above exhibit
	very different answers to the question of why it is 
	important to obtain good scene descriptions.
In traditional computer vision, 
	a good scene description is of interest for qualitative, humanistic reasons. 
This motivation makes it very difficult to evaluate methods, 
	since human input is required to determine the quality of a result. 
In contrast, in compression-based vision research, 
	a good scene description is of interest for quantitative, mechanistic reasons. 
This takes the human out of the evaluation loop, 
	making it much easier to compare techniques.

\subsubsection{Stereo Correspondence}

The previous section showed that it was possible to reformulate
	a very abstract version of the vision problem in terms of compression.
This and the following two sections show how this reformulation can work
	for \textit{specific} vision problems.
But for the case of the stereo correspondence problem,
	the argument has already been made,
	by Mumford (emphasis in original):
	
\begin{quote}
I'd like to give a more elaborate example to show how MDL can lead you to
	the correct variables with which to describe the world using an old and familiar
	vision problem: the stereo correspondence problem. 
The usual approach to stereo vision is to apply our knowledge of the three-dimensional structure of the
	world to show how matching the images $I_{L}$ and $I_{R}$ from the left and right eyes
	leads us to a reconstruction of depth through the ``disparity function'' $d(x,y)$
	such that $I_{L}(x+d(x,y),y)$ is approximately equal to $I_{R}(x,y)$. 
In doing so, most algorithms take into account the ``constraint'' that most surfaces in the world are
	smooth, so that depth and disparity vary slowly as we scan across an image.
The MDL approach is quite different. 
Firstly, the raw perceptual signal comes as two sets of $N$ pixel values $I_{L}(x,y)$ and $I_{R}(x,y)$ 
	each encoded up to some fixed accuracy by $d$ bits, totaling $2 d N$ bits. 
But the attentive encoder notices how often pieces of the left image code nearly 
	duplicate pieces of the right code: this is a common pattern that 
	cries out for use in shrinking the code length. 
So we are led to code the signal in three pieces:~first the raw left image $I_{L}(x,y)$; 
	then the disparity $d(x,y)$; and finally the residual $I_{R}(x, y)$.
The disparity and the residual are both quite small, 
	so instead of $d$ bits, these may need only a small number $e$ and $f$
	bits respectively.
Provided $d > e + f$, we have saved bits.
In fact, if we use the constraint that surfaces are mostly smooth,
	so that $d(x,y)$ varies slowly, we can further encode $d(x,y)$ 
	by its average value $d_{0}(y)$ on each horizontal line
	and its $x$-derivative $d_{x}(x,y)$ which is mostly much smaller.
The important point is that MDL coding leads you to introduce the third coordinate of space, 
	i.e. to discover three-dimensional space!
A further study of the discontinuities in $d$, and the ``non-matching'' pixels
	visible to one eye only goes further and leads you to \textit{invent a description}
	of the image containing labels for distinct objects,
	i.e. to \textit{discover that the world is usually made up of discrete objects}
	~\cite{Mumford:1994}.
\end{quote}

Note how a single principle (compression)
	leads to the rediscovery of structure in visual reality
	that is otherwise taken for granted
	(authors of object recognition papers do not typically feel obligated
	to justify the assumption that the world is made up of discrete objects).
Mumford's thought experiment also emphasizes the intrinsic scalability 
	of the compression problem:
	first one discovers the third dimension, 
	and then that the world is made up of discrete objects.

\subsubsection{Optical Flow Estimation}

Another traditional task in computer vision is optical flow estimation.
This is the problem of finding the apparent motion 
	of the brightness patterns in an image sequence.
The optical flow problem can be reformulated as a specialized compression technique
	that works as follows.
Consider a high frame rate image sequence (say, 100 Hz).
Because of the high frame rate,
	the scene does not change much between frames.
Thus, a good way to save bits would be to encode and transmit full frames
	at a lower rate (say, 25 Hz), 
	and use an interpolation scheme to predict the intermediate frames.
The predicted pixel value would then be used as the mean for the 
	distribution used to encode the real value,
	and assuming the predictions were good, substantial bit savings
	would be achieved while maintaining losslessness.
Now, a ``dumb'' interpolation scheme could just linearly interpolate
	a pixel value by using the start and end frames.
But a smarter technique would be to infer the motion of the pixels
	(i.e. the optical flow) and use this information to do the interpolation.
	
The simplest encoding distribution to use would be a Gaussian with unit variance
	and mean equal to the predicted value.
In that case, the codelength required to encode a pixel
	would be simply the squared difference between the prediction
	and the real outcome, 
	plus a constant corresponding to the normalization factor.
A smarter scheme 
	might take into account the local intensity variation -
	if the intensity gradient is large, 
	it is likely that the prediction will be inaccurate,
	and so a larger variance should be used for the encoding distribution.
The resulting codelength for a single interpolated frame would be:

\begin{equation}
\sum_{x, y} \frac{(I(x,y) - I_{GT}(x,y))^{2}}{||\nabla I_{GT}(x,y)||^{2} + \epsilon}  + k(I_{GT}(x,y))
\end{equation}
	
Where $I(x,y)$ is the real frame and $I_{GT}(x,y)$
	is the image predicted from the optical flow,
	and the $k(\cdot)$ is a term corresponding to the normalization factor.
Since shorter codelengths are achieved by 
	improving the accuracy with which $I_{GT}$ predicts $I$,
	this shows that improvements in the optical flow estimation
	algorithm will lead to improvements in the compression rate.
Indeed, with the exception of the $k(\cdot)$ terms,
	the above expression is equivalent to an evaluation metric
	for optical flow algorithms proposed by~\cite{Baker:2007}
	(compare their Equation 1 to the above expression).
The compression-equivalent scheme is much simpler than the other metric 
	proposed by~\cite{Baker:2007},
	which involves the use of a complicated experimental apparatus 
	to obtain ground truth.
The compression metric permits
	\textit{any} image sequence to be used as empirical data.
		
\subsubsection{Segmentation}

The task of image segmentation has been a subject of research in 
	computer vision for more than thirty years. 
A major obstacle in this research is that there is no 
	single, definitive articulation of what the goal of 
	an image segmentation algorithm should be. 
The rough outlines of the problem are widely agreed upon: 
	a segmentation algorithm should partition the image 
	into a small number of simple, homogeneous regions. 
The difficulty is in the precise definition of the words
	``small'', ``simple'', and ``homogeneous'', 
	regarding which there is no widespread agreement. 

The segmentation problem can be formulated as a compression problem as follows. 
A special compressor is used that represents an image as a set of regions. 
The pixels of each region are encoded using a specialized statistical model.
Using a region-specific model, instead of a generic model for the entire image, 
	can in principle help to reduce the overall codelength. 
However, several conditions must be met for this to work. 
First, the pixels assigned to a region must be very similar to one another. 
Second, the format requires that the contours of the regions must also be encoded, 
	so the region boundaries must be simple. 
Finally, encoding the specialized model requires an overhead cost for each region, 
	so the total number of regions should be kept to a minimum. 
These three considerations supply cleanly justified definitions
	for the problematic words (homogeneous, simple, small) mentioned above. 

The MDL/compression approach to segmentation has been followed by 
	several authors~\cite{Leclerc:1989,Kanungo:1994}. 
The following is a brief discussion of a method proposed by Zhang and Yuille~\cite{Zhu:1996}. 
In this paper the segmentation problem is formulated as a minimization of the functional:

\begin{eqnarray}
\sum_{i}^{M} \left\{\frac{\mu}{2} \int_{\partial R_{i}} ds - 
	\log P \left( {I_{x,y}:(x,y) \in R_{i} }|\alpha_{i} \right) + \lambda \right\}
\label{eq:zhumdl}
\end{eqnarray}

This functional is a sum over segmented regions.
There is an cost associated with the region boundary (contour integral),
	a cost resulting from encoding a set of pixels given a
	particular region model ($\log P$ term),
	and a constant cost for encoding a region model ($\lambda$).
The goal is to find a good set of region boundaries $\partial R_{i}$
	and associated region model parameters $\alpha_{i}$
	that minimize the sum.
This illustrates a competition between the need 
	to package similar pixels together so that narrow, region-specific
	model distributions will describe them well,
	and the need to use a small number of regions with simple boundaries.
	
Note that the focus of the paper~\cite{Zhu:1996}
	is on the development of an algorithm  
	for finding a good minimum of Equation~\ref{eq:zhumdl}.
Little effort is spent on finding good region models  
	or efficient boundary encoding methods.
The paper reports only segmentation results, 
	not compression results.
This is because the compression idea is viewed simply as a trick
	that allows good segmentations to be obtained.
The purpose of this paper, of course,
	is to advocate the opposite approach.
	
\subsubsection{Face Detection and Modeling}
\label{sec:facedetmod}

Imagine that the target $T$ used in the CRM 
	is the image database hosted on the popular 
	internet social networking site Facebook.
This enormous database contains many images of faces. 

Faces have a very consistent structure.
There is a significant literature on modeling faces~\cite{Blanz:2003,Decarlo:1998},
	and several techniques exist that can produce convincing
	reproductions of face images from models with a
	small number of parameters.
Given a starting language $L$,
	by adding this kind of model based face rendering technique
	a new language $L_{f}$ can be defined that contains the ability 
	to describe scenes using face elements.
Since the number of model parameters required is generally small
	and the reconstructions are quite accurate,
	it should be possible to significantly compress the Facebook
	database by encoding face elements instead of raw pixels when appropriate.

However it is not enough just to add face components 
	to the description language.
In order to take advantage of the new face components of the language 
	to achieve compression,
	it is also necessary to be able to obtain good descriptions
	$D_{L_{f}}$ of images that contain faces.
If unlimited computational power were available,
	then it would be possible to test each image subwindow to determine
	if it could be more efficiently encoded by using the face model.
But the procedure of extracting good parameters for the face model is relatively expensive,
	so this brute force procedure is inefficient.
A better scheme would be to use a fast classifier for face detection~\cite{Viola:2004}.
The detector scans each subwindow,
	and if it reports that a face is present,
	the subwindow is encoded using the face model component.
Bits are saved only when the detector correctly predicts that
	the face-based encoder can be used to save bits for the subwindow.
A false negative is a missed opportunity,
	while a false positive incurs a cost related to the inappropriate
	use of the face model to encode a subwindow.
In other words, the face model implicitly defines a virtual label for each subwindow,
	which is true if the subwindow can be encoded more efficiently using the face model.
This implies that a vast but completely unlabeled image database 
	can now be used to evaluate the performance of a face detection system.

\section{Conclusion}

Empirical sciences,
	such as physics and chemistry,
	have produced some of the most remarkable successes in the intellectual history
	of humankind.
Empirical sciences proceed by formulating theories about reality,
	and then testing, refining, and sometimes falsifying those
	theories by comparing their predictions to experimental results.
This process, when rigorously applied, can lead to rapid progress.
Under the current paradigm of computer vision,
	new ideas take the form of computational methods or mathematical theorems,
	not hypotheses about physical reality.
This means that it is difficult to decisively compare ideas,
	and so it is hard to make progress.
	
This paper proposed a way to reformulate computer vision as an empirical science.
The key obstacle is that the traditional scientific method can
	not be productively applied to answer the questions of interest in the field.
The contribution of this paper is a refined version of the scientific method
	in which vast databases are used as the empirical component
	instead of the results of controlled experiments.
By applying the Compression Rate Method to large databases of natural images,
	it should be possible to develop empirical theories of visual reality.
	
The connection to empirical science comes from the No Free Lunch theorem of data compression,
	which shows that compression can be achieved for some images
	only at the price of inflating other images.
If a precise computational definition of ``natural images'' were available, 
	then one could instruct the compressor to shrink those images,
	while inflating non-natural images.
This definition is not available,
	but the quality of a proposal definition can be evaluated
	by measuring the codelength it achieves on a large database.
Thus, an increasingly precise characterization of natural images can be developed
	by iterative refinement of a series of proposal definitions.
	
The compression principle provides new answers to two long-standing
	issues in the philosophy of science:
	the problem of induction and the problem of demarcation.
The problem of demarcation is solved by accepting a theory as scientific
	if and only if it can be used to achieve compression.
Since compression requires that a theory reassign probability away from
	some outcomes and toward other outcomes,
	this is simply a graduated version of Popper's principle of falsifiability.
The problem of induction is solved by using ``vast'' databases
	that have Kolmogorov complexities that are much larger
	than the characteristic lengths of Turing machine simulator programs.
In this ``vast data regime'', 
	statistical inference achieves intersubjective verifiability.
	
The second part of the paper discusses specific ideas related 
	to the application of the Compression Rate Method 
	to the field of computer vision.
The section begins with a critical summary of current evaluation 
	methods in computer vision.
These methods suffer from a number of practical and conceptual difficulties.
One simple issue is that obtaining ground truth data
	requires a lot of human effort.
An important conceptual issue is that for many computer vision tasks,
	such as image segmentation and edge detection,
	there is no precisely defined ``correct'' answer.
More abstractly, the current evaluation paradigm has bad scaling properties,
	since each task requires its own evaluator.

The vision-as-compression approach yields many advantages.
The approach provides a very strong degree of methodological rigor,
	since new results can easily be checked by a third party.
It is also a much more scalable evaluation scheme,
	since multiple methods can be evaluated using the same database.
By using large databases, 
	it becomes possible to justify the construction of highly complex theories.
Finally, the approach should allows the field to 
	decisively compare competing theories
	and thus make systematic progress.

A crucial argument was that the concerns of image compression and 
	computer vision are deeply related:
	most vision tasks can be reformulated as specialized compression techniques.
The paper discussed stereo matching, image segmentation,
	optical flow estimation, and face detection/modeling,
	but similar arguments can be made for other tasks,
	such as egomotion inference and 3D reconstruction.
The paper also provides an abstract formulation 
	of computer vision as image compression.
The point here was that a scene description is valuable for 
	quantitative, mechanistic reasons (it allows compression)
	instead of the qualitative, humanistic reasons.

\bibliographystyle{IEEEtran}
\bibliography{IEEEabrv,../latex/bibtex/allrefs}

\end{document}